\documentclass[a4paper]{cas-dc}

\ExplSyntaxOn

\cs_set:Npn \__make_tbl_caption:nn #1#2
{
  \l_tbl_align_tl
  \skip_vertical:N \l_tbl_abovecap_skip
  {
    \parbox{\l_tbl_width_dim}
    {
      \sffamily\small
      \textbf{#1}\ #2
      \par
      \vskip 4pt
    }
  }
  \skip_vertical:N \l_tbl_belowcap_skip
}

\ExplSyntaxOff

\usepackage[authoryear,longnamesfirst]{natbib}
\usepackage{float}
\usepackage{orcidlink}

\def\tsc#1{\csdef{#1}{\textsc{\lowercase{#1}}\xspace}}
\tsc{WGM}
\tsc{QE}

\begin{document}
\let\WriteBookmarks\relax
\def\floatpagepagefraction{1}
\def\textpagefraction{.001}

\shorttitle{From Tweets to Trades: Analyzing the Influence of Public Mood over Stock Market Performance in Turkiye}

\title [mode = title]{From Tweets to Trades: Analyzing the Influence of Public Mood over Stock Market Performance in Turkiye}

\author[1]{Ece Elif Adak\orcidlink{0009-0008-2624-718X}}
\ead{ece.adak@std.bogazici.edu.tr}

\author[2]{Berta\c{c} \c{S}akir \c{S}ahin\orcidlink{0000-0003-0414-5402}}
\ead{bertacsa@yildiz.edu.tr}

\author[1]{\c{S}aziye Bet\"{u}l \"{O}zate\c{s}\orcidlink{0000-0003-3254-0960}}
\ead{saziye.ozates@bogazici.edu.tr}
\cormark[1]

\address[1]{Boazici University, 34342 Bebek/Istanbul, Turkiye}
\address[2]{Y{\i}ld{\i}z Technical University, 34220 Esenler/Istanbul, Turkiye}

\cortext[1]{Corresponding author: Şaziye Betül Özateş}

\begin{abstract}
\textbf{\bf Purpose:} This study examines whether domain-specific public mood is associated with stock-market dynamics and whether these relationships vary across communication domains and market conditions. It distinguishes public mood from investor sentiment and investigates whether heterogeneous sources of public communication exhibit different relationships with market behaviour.
\\
\textbf{Design:} The study analyses 610,422 posts published by 176 curated X accounts between January 2022 and December 2023, covering Politics and Government, Economy and Finance, and Media and Society. Posts are classified using fine-tuned Turkish transformer models under three domain-specific and one pooled regime. Public mood measures are constructed at daily, weekly, and monthly frequencies and examined alongside BIST100 and BIST30 market measures using correlation, Granger causality, vector autoregression, and impulse response analyses across the full period and selected market conditions.
\\
\textbf{Findings:} Public mood is not associated with the direction of stock-market returns but is associated with the magnitude of price movements, particularly for Media and Society and pooled communication. These relationships become stronger at longer aggregation frequencies. Predictive relationships are concentrated in Economy and Finance communication, while their magnitude and direction vary across market conditions, particularly during the 2023 election period. The pooled measure largely reflects the most active communication domain.
\\
\textbf{Originality:} The study contributes to behavioral-finance research by incorporating communication - domain heterogeneity into the analysis of public mood and market dynamics. It also demonstrates how aggregating heterogeneous sources can obscure domain-specific relationships between public communication and financial markets.

\end{abstract}


\maketitle

\section{Introduction}
\label{sec:introduction}

Financial markets face a continuous flow of social-media commentary that rapidly spreads political views, economic forecasts, news and emotional narratives beyond official disclosures. Under the efficient market hypothesis, prices should quickly incorporate value-relevant public information, leaving historical linguistic tone with little explanatory power once market data are considered \citep{fama1970}, while recent evidence supports semistrong efficiency over longer horizons \citep{agrrawal2025}. Behavioural finance, however, suggests that beliefs, attention and non-fundamental optimism or pessimism can influence trading and valuation when uncertainty is high and arbitrage is limited \citep{baker2006}. 

We therefore distinguish public mood the positive, neutral or negative linguistic valence measured separately in political, economic and social communication from investor sentiment. Since accounts are selected by speaker rather than topic, posts need not originate from investors or concern financial markets; interpreting their tone as investor sentiment would attribute identities and market-specific beliefs that the texts do not establish. Public mood instead captures an observable feature of communication \citep{dean2017}, consistent with textual finance research distinguishing public tone from investors’ latent sentiment \citep{kearney2014,loughran2016}. This mood may influence markets through attention, information diffusion, repeated narratives and interpretations of uncertainty, or reflect beliefs formed elsewhere, making the relationship potentially bidirectional. Evidence from X links online communication to returns, trading volume and volatility \citep{bollen2011,sul2017}.
 
Turkiye provides a useful setting. Direct participation in Borsa Istanbul rose sharply after 2020. By the end of May 2024 more than 8.2 million investors held Borsa Istanbul equities, with roughly 5.5 million holding BIST100 constituents \citep{cmb2024,mki2024}. A larger investor base increases the importance of the public channels through which economic and political narratives reach market participants. The period from January 2022 to December 2023 combines persistent inflation and exchange-rate pressure with several discrete events. The February 2023 earthquakes, the May 2023 presidential and parliamentary elections, and a pronounced wave of initial public offerings in the second half of 2023. These produce economically distinct conditions under which the relationship between public communication and market behaviour may strengthen, weaken or change direction.

Research on textual information and Turkish markets has advanced quickly. These studies establish that user-generated online text can carry financial information, that the relationship may be bidirectional, and that its intensity varies. What remains open is not whether social-media text relates to market outcomes but which tone of communication is being measured and whether different domains of communication behave alike. Current evidence remains disjointed on three interconnected inquiries. First, social-media tone is frequently consolidated into a single metric, although political, economic and social discussions may convey distinct information and may trigger various channels of attention and risk awareness. Second, research that accurately gauges investor sentiment often depends on messages created by investors or specific to finance; broader samples from influential public accounts necessitate a framework that does not assume an investor’s identity. Third, temporal variation is usually studied separately from domain composition, leaving limited evidence on whether the direction of information transmission itself changes across normal, crisis and political uncertainty periods. These issues are especially relevant in an emerging market in which political developments, macroeconomic conditions and social events can influence market expectations
\citep{kilimci2020,cagli2020,liu2023,liu2026}.
 
To address them, we analyse 610,422 posts published by 176 influential X accounts between January 2022 and December 2023. We group the accounts into Politics and Government, Economy and Finance, and Media and Society. We fine-tune seven transformer-based language models on novel, manually labelled Turkish data for three-class sentiment classification and use the best-performing configuration for each dataset to construct political, economic, social and combined sentiment score calculation. We align these with the BIST100 and BIST30 at daily, weekly and monthly frequencies. We then use Augmented Dickey--Fuller tests, Pearson and Spearman correlations, bidirectional Granger causality tests, vector autoregressions and impulse response functions to examine associations with prices, returns, trading activity and volatility. We repeat the analysis within the 2022 baseline, the 2023 earthquake quarter, the election quarter and the post-election IPO period, and again at the level of individual accounts.
 
 
We make four contributions. First, we define the construct under examination clearly. By characterising the textual variable as domain-specific public mood, we separate the observable valence of influential public communication from investor sentiment in the narrower sense used in behavioural finance. Second, we document variation across domains (political, economic and social), moods are not interchangeable. Their strongest associations arise for different market outcomes and different evaluative criteria. Third, we combine domain variation with bidirectional transmission and changing market conditions, showing that not just the intensity but also the direction the sentiment--market relationship shift when political or disaster-driven uncertainties arise. Fourth, we provide a substantial Turkish-language dataset and a systematic comparison of seven transformer models for three-class sentiment classification, establishing a reproducible framework for further work on public communication and emerging financial markets. We treat the evaluation of language models as a measurement contribution. The financial contribution concerns how different types of public tone relate to, and are related to by the market under varying conditions.
 

\section{Theoretical Background and Hypotheses}
\label{sec:theory}
 
The relevance of public communication to finance can be framed against informational efficiency. In a semi-strong efficient market, publicly available information should be reflected in prices quickly. So historical linguistic tone should add little once market data are taken into account \citep{agrrawal2025,fama1970}. Behavioral finance allows a different outcome. Where valuation is uncertain and arbitrage is limited; beliefs, attention and swings of optimism or pessimism can affect trading and asset prices even when they do not correspond to fundamentals \citep{baker2006}. Noise trader theory gives a specific mechanism. Investors may trade on unclear or poorly grounded signals while rational arbitrageurs act cautiously because mispricing can worsen before it corrects \citep{delong1990}. Public mood can therefore relate to prices and volatility by spreading information and by leading to correlated trading not based on fundamentals.

We use public mood to refer to the positive, neutral or negative sentiment that can be observed in social-media communication from influential sources. Public mood captures positive, neutral or negative tone in influential social media communication, rather than investors’ beliefs. Social media spreads narratives rapidly through networks, although investors differ in access, interpretation and responsiveness. Salience attracts attention \citep{barber2008}, while uneven information absorption \citep{hong1999} may yield clearer effects over longer windows, without new information. Political mood concerns policy continuity, institutional credibility and regulation; economic mood concerns inflation, rates, currencies, growth and cash flows; social mood may affect trading or volatility more than returns. Elections and policy changes can raise risk premiums, particularly in fragile economies \citep{pastor2013}. Political optimism may signal strategic communication or become a contrary indicator when expectations are priced in. Aggregation can therefore obscure distinct or offsetting effects.
 
The direction of transmission is also unlikely to run only from communication to the market. Public mood may influence prices by altering attention, expectations and perceived uncertainty. But gains, losses and volatility can reshape subsequent public discussion. Evidence from social media points to a reciprocal relationship between online sentiment and stock prices \citep{liu2023}, while research on Borsa Istanbul shows that predictive sequence between sentiment proxies and excess returns emerges only in particular episodes \citep{cagli2020}. The resulting process is better described as feedback than as a one-way effect. As such, we interpret bidirectional Granger tests as tests of predictive precedence, not as evidence of structural economic causality.
 
The Adaptive Markets Hypothesis links these arguments by treating market efficiency and investor behaviour as outcomes that evolve with the environment \citep{lo2004}. A disaster, an election or a sharp change in market participation can alter the credibility of public messages, the composition of active investors and the decision rules they use. A relationship observed in a stable period may weaken, disappear or reverse when uncertainty dominates. The integrated framework consequently treats the public mood--market relationship as domain-specific, bidirectional and state-dependent, and provides the basis for comparing political, economic, social and combined mood across the baseline, earthquake, election and post-election periods. We test the following hypotheses.
 
\begin{itemize}
    \item \textbf{H1.}
Domain-specific public mood contains information relevant to Borsa Istanbul market outcomes.
\item \textbf{H2.}
Political, economic and social public mood exhibit significantly different relationships with market returns, trading activity and volatility.
\item \textbf{H3.}
As textual and market data are aggregated over longer horizons, the relationship between public mood and market outcomes becomes more pronounced.
\item \textbf{H4.}
Predictive precedence runs in both directions between public mood and market outcomes.
\item \textbf{H5.}
The magnitude, direction and sign of the relationship vary across normal, disaster-related, politically uncertain and post-election regimes.
\end{itemize}

\section{Literature Review}
\label{sec:literature}
 
\subsection{Public mood, social media tone and market outcomes}
\label{subsec:publicmood}
 
Sentiment analysis is widely used to examine how public views relate to financial markets, but terminology varies across media tone, textual sentiment, investor opinion, social-media sentiment, and collective mood. These concepts overlap but do not necessarily measure the same underlying construct. Results depend on the information source, classification method, and outcome examined \citep{kearney2014}, while the meaning of positive and negative language is context-dependent and tied to the documents and actors producing it \citep{loughran2016}. Thus, linguistic valence in public communication should not automatically be interpreted as investor beliefs.
 
That publicly available language carries market-relevant information is well established. \citet{antweiler2004} found by analysing internet stock message boards that online discussion is more informative about volatility than about large return movements. \citet{tetlock2007} reached a comparable conclusion using financial newspaper content. Media pessimism is followed by downward price pressure and a subsequent reversal, and unusually high or low pessimism is associated with elevated trading volume. These results suggest that textual tone may reflect temporary price pressure, shifts in attention or uncertainty rather than durable changes in fundamental value, and that returns are only one metric through which the market may respond.
 
The growth of social media extended this work beyond financial journalism and investor forums. Rather than building mood indicators from messages about stocks alone, \citet{bollen2011} derived them from a broad Twitter stream and found that some dimensions of public mood help predict movements in the Dow Jones Industrial Average while others relate much less. Collective public expression is therefore multidimensional, and not all of its emotional components carry the same financial weight. By contrast, \citet{li2018} examined stock-specific microblogs and found their information content associated with abnormal returns, trading volume and volatility, with these relationships strengthening when messages are weighted by the social influence of the users producing them. Taken together, this evidence indicates that the source and scope of online communication matter. A broad public-mood indicator and opinion expressed within an investment community may both relate to markets without measuring the same phenomenon.
 
The informational content of social-media language also depends on the wider information environment. \citet{khan2022} showed that traditional news coverage and social-media coverage are associated with different patterns of turnover and volatility. \citet{ren2024} showed how social platforms reshape the consumption of traditional media news. Social media may repeat narratives available elsewhere, but repetition can still influence attention and trading. The predictive value of textual tone therefore cannot be assessed without considering who produces the content, whether that producer has a financial focus, and how the information reaches market participants.
 
Prior research thus provides substantial evidence that public language is associated with returns, trading volume and volatility, and much weaker support for treating every textual indicator as a uniform measure of investor sentiment. Measures constructed from financial forums, firm-specific messages, professional news and broad social-media communication reflect different authors and different information. Much of this literature also aggregates positive and negative communication into a single market-wide signal even though political, economic and social discourse may differ in credibility, salience and financial relevance. This motivates our stratification of public mood by domain.
 
\subsection{Social media analytics and evidence from Borsa Istanbul}
\label{subsec:bistevidence}
 
Research on social media and Borsa Istanbul has developed along two connected lines: constructing reliable sentiment measures from Turkish-language content, and testing whether those measures inform market outcomes. Early work targeted finance-related messages. \citet{kilimci2020} paired three word-embedding models, Word2Vec, GloVe and FastText, with three deep-learning architectures, convolutional and recurrent neural networks and long short-term memory networks, to predict the direction of nine actively traded banking stocks using Twitter, financial news sites and public disclosures, and report considerable variation across text sources, indicating that predictive performance depends on the information source and not only on the classification approach. \citet{ates2021} subsequently linked sentiment derived from Turkish financial tweets to daily BIST30 returns using Pearson correlation and Granger causality. Their Granger results are directly relevant here. Over a long sample the predictive sequence runs from returns to sentiment, while some sentiment measures precede returns only within a politically eventful short window covering the June 2018 election. Direction, in other words, was already known not to be fixed on Turkish data.
 
A second line shows that the financial relevance of social-media content cannot be inferred from its availability alone. \citet{ismayil2023}, focused on two airlines listed on Borsa Istanbul and separated posts concerning the companies' shares from posts about their products and services. They found that Twitter activity concerning the shares positively related to stock performance while broader discussion of the companies did not. A relatively small number of financially relevant messages produced correlations comparable to those from the larger set of share-related posts. Which indicates that filtering for market relevance matters more than increasing the volume of text. \citet{cam2024} concentrated on the classification stage, labelling Turkish posts gathered through Borsa Istanbul-related hashtags with a multilingual sentiment lexicon. Then compared six supervised classifiers trained on those labels, naive Bayes, logistic regression, support vector machines, \(k\)-nearest neighbours, decision trees and a multilayer perceptron, of which the support vector machine and the perceptron performed best. That work advanced the measurement of Turkish financial sentiment, but high classification accuracy does not imply that the resulting indicator will predict returns, volume or volatility. The distinction between successful text classification and successful market prediction is essential when comparing evidence across studies.
 
Work published in Borsa Istanbul Review has moved beyond positive--negative analysis. \citet{akdogan2024} applied a gated recurrent unit classifier to identify posts concerning Borsa Istanbul, labeled their sentiment with a fine-tuned Turkish BERT model. They used the resulting social, cognitive and behavioural features in linear and Lasso regressions, random forests and gradient boosting to model the opening level, trading volume and volatility of the BIST100. Their results showed that social-media information may relate to several dimensions of market activity, while also stressing the need to separate financially relevant content from the broader stream before constructing explanatory variables. \citet{ibrahim2025} extended the comparison across information sources by combining local social-media posts with domestic and international news narratives. They found that using sentiment features fed into gradient boosting, XGBoost and random forest models with SHAP values attributing the predictions. International news sources generally played a larger role in forecasting the Turkish stock market than local Twitter content. The informational value of text appears to depend on its source, reach and connection to the economic environment rather than on its origin in social media as such.
 
Other evidence emphasises volatility and changing market conditions. \citet{bozma2020} used sentiment from company-specific tweets within a multivariate GARCH framework and documented relationships between Twitter-based measures and volatility transmission among selected Borsa Istanbul stocks. \citet{sevinc2026} examined social-media and investor-forum posts concerning stocks identified by the Capital Markets Board in connection with manipulation, using a lexical framework that added a suspicious content category to conventional sentiment classes. The effects of online communication differed across bearish, normal and bullish conditions and across levels of trading activity.
 
Taken together, the Turkish evidence confirms that social-media information can be associated with market direction, returns, trading activity and volatility, and reveals substantial differences in textual sources, filtering procedures and evaluation criteria. Most studies rely on finance-specific hashtags, firm-related messages, investor forums or content screened for market relevance. Thus differ conceptually from an indicator constructed from political actors, public institutions, financial commentators, news organisations and other influential accounts. Previous work also generally transforms textual information into an aggregate sentiment measure, even though distinct areas of public communication may carry different market-relevant information.
 
\subsection{Domain heterogeneity, bidirectional transmission and state dependence}
\label{subsec:domainhet}
 
Textual information should not be treated as a single uniform market signal, since the financial importance of language can depend on the topic and on the type of risk or expectation attached to it. \citet{calomiris2019} used news from 51 developed and emerging economies to construct topic-specific measures of sentiment, news flow and unusual language. They found that these hold information about future returns, volatility and drawdowns with predictive value differing across countries and horizons. Combining topics into one tone indicator can therefore conceal distinct economic signals.
 
The relationship between textual indicators and market outcomes may also involve feedback rather than one-way transmission. \citet{liu2023} reported a generally positive link between social-media investor sentiment and stock prices while noting that this relationship can weaken, vanish or reverse in particular periods. Intraday evidence suggested that sentiment may carry forward-looking information even as price movements remain closely tied to subsequent online sentiment. \citet{liu2026} extended this to the intraday microstructure. Evidence from Borsa Istanbul points the same way. \citet{cagli2020} found no overall causality between a composite investor-sentiment index and BIST100 returns using standard full-sample Granger tests. But distinct causality episodes appear once nonlinearities and structural change are assessed with a recursive evolving-window method, with predictive influence running from sentiment to the market, from the market to sentiment, or in both directions depending on the proxy and period. Constant-parameter estimates may conceal relationships that exist only during specific episodes.
 
The predictive role of sentiment also depends on the economic environment. \citet{chung2012} showed that investor sentiment predicts several stock portfolios during expansions while its predictive ability is generally weak during recessions. This implies that the effects of optimism, misvaluation and limits to arbitrage vary with conditions. The literature thus suggests that textual information may be topic-specific, linked to markets through feedback, and dependent on the prevailing regime. It remains unclear whether political, economic and social public mood exhibit different predictive patterns and whether these relationships shift jointly across normal, crisis and politically uncertain periods. We address that gap by examining domain-specific public mood, bidirectional predictive precedence and state dependence within a common empirical framework.

\section{Methodology}
\label{sec:data}
 
\begin{figure*}
\centering
\includegraphics[width=\textwidth]{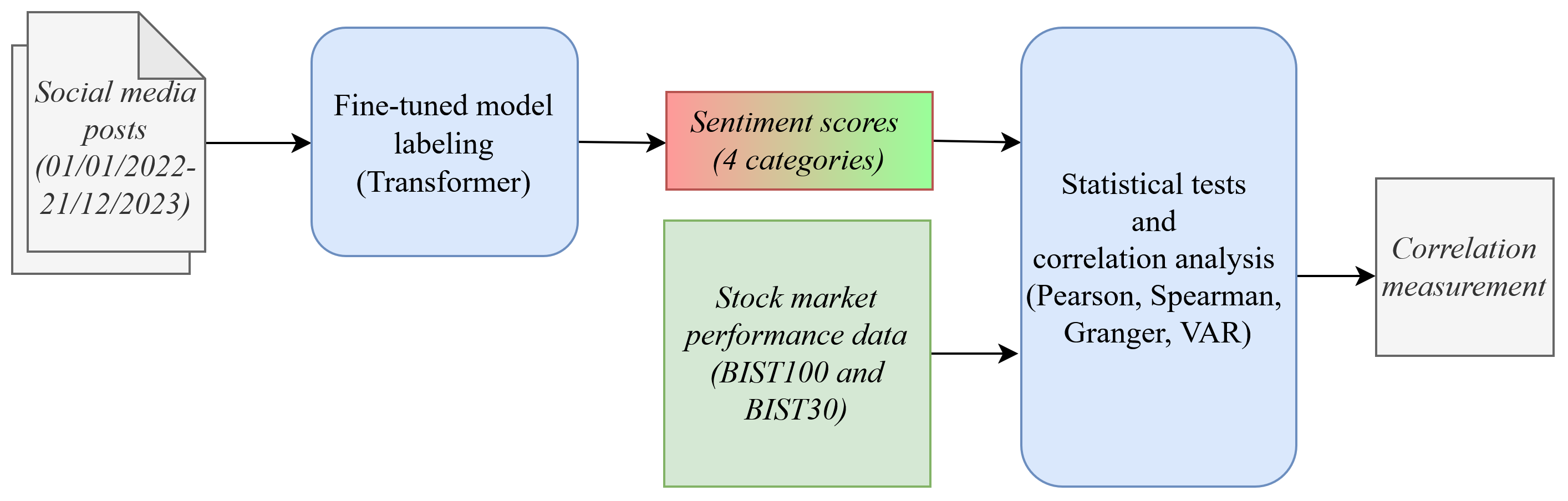}
\caption{Methodological outline of the study.}
\label{fig:method}
\end{figure*}

We collect and analyze 600K posts published by 176 curated X accounts between January 2022 and December 2023. The accounts are stratified into three broad domains: Politics and Government, Economy and Finance, and Media and Society. The posts are classified using four fine-tuned Turkish transformer models under three domain-specific training regimes and one pooled regime. To examine the relationship between social media activity and financial market performance, we aggregate the resulting predictions into daily, weekly, and monthly time series and evaluate their relationships with BIST100 and BIST30 performance data using different statistical testing approaches.
Figure~\ref{fig:method} illustrates the overall methodology, and the following subsections describe each step in detail.

\subsection{Corpus construction}
\label{subsec:corpus}
 
The corpus is selected and constructed by speaker rather than by topic. Turkish studies of social media and Borsa Istanbul have collected text through financial hashtags, cashtags, firm names or disclosure feeds. Under such a filter the sentiment being measured is by construction sentiment about the market. Removing the filter permits a different question: whether discourse that is not about the market is still relevant to it. Answering that question requires selection criteria that operate on accounts rather than on posts.
 
We assigned accounts to categories before collecting any posts. Politics and Government contains the official accounts of the Presidency and the ministries, the President of Turkiye, the ministers serving during the study window, the Speaker of the Grand National Assembly, and the leaders of the main opposition parties represented in parliament. Economy and Finance contains the Central Bank of the Republic of Turkiye and its governors during the window, the official accounts of Borsa Istanbul and of the Capital Markets Board, the economic ministries, and financial news outlets, economists and market commentators. Media and Society contains general news outlets, journalists, selected public figures, and the official accounts of the five largest municipalities. Our assignment is not strictly disjoint, several ministries and ministers whose remit spans economic policy and public services belong to more than one category, so category post counts sum to more than the number of distinct posts.
 
Two criteria governed our inclusion decisions. Institutional and office-holding accounts entered on the basis of their role, irrespective of audience size. All remaining accounts were required to exceed one million followers. The follower condition is a reach condition, a relationship between public discourse and an aggregate index is plausible only for discourse reaching enough market participants to matter. It also has a cost, stated in Section~\ref{subsec:limitations}. It removes the retail conversation, which is where investor sentiment would be found.
 
The corpus comprises 610,422 distinct posts published by 176 accounts between 1 January 2022 and 31 December 2023. 119,877 posts were published by Politics and Government accounts, 256,135 by Economy and Finance accounts and 297,857 by Media and Society accounts. Posts published on weekends and market holidays are assigned to the next trading day, on the assumption that discourse produced while the exchange is closed can only be acted upon at the next open. This added up to 500 trading days over the two year period.
 
\subsection{Annotation}
\label{subsec:annotation}
 
We drew three labelled training sets from the corpus, one per category, each restricted to posts by accounts only in that category, and a fourth pooling all three. Human annotators labelled posts as positive, negative or neutral. We excluded posts too vague in sentiment to assign and removed highly similar posts so that learning achieved by the models were not diminished. Table~\ref{tab:datasets} reports the resulting sizes and class distributions.
 
\begin{table*}[
    width=.9\linewidth,
    cols=7,
    pos=h,
    align=\centering
]
\caption{Labelled dataset sizes and class distribution.}
\label{tab:datasets}
\footnotesize
\setlength{\tabcolsep}{4pt}
\begin{tabular}{lrrrrrrr}
\hline
Dataset & Total & Train & Valid. & Test & Positive & Negative & Neutral \\
\hline
Politics and Government & 1,500 & 1,000 & 200 & 300 & 959 & 460 & 81 \\
Economy and Finance & 1,506 & 1,006 & 200 & 300 & 705 & 652 & 149 \\
Media and Society & 1,502 & 1,002 & 200 & 300 & 764 & 599 & 139 \\
Combined & 4,508 & 3,008 & 600 & 900 & 2,428 & 1,711 & 369 \\
\hline
\end{tabular}
\end{table*}
 
We assessed inter-annotator agreement on 150 posts drawn from all three categories and labelled independently by two annotators. Weighted Cohen's \(\kappa\) across the three classes reached 0.817, which falls in the highest band of the conventional interpretation \citep{landis1977}. Two features of the labelled data bear on interpretation. Excluding vague posts removes precisely the cases a classifier finds hardest, so test-set metrics may overstate performance on the full corpus. And the positive class is the majority in all three category sets, especially Politics and Government, where 959 of 1,500 posts are positive and only 81 are neutral.
 
\subsection{Sentiment classification}
\label{subsec:classification}
 
We fine-tuned seven pre-trained transformer models on each of the four labelled datasets for three-class sentiment classification: mBERT \citep{devlin2019}, BERTurk, DistilBERTurk, BERT5urk \citep{schweter2025}, and ConvBERTurk, which applies the ConvBERT architecture of \citet{jiang2020}, TurkishBERTweet \citep{najafi2024} and Gemma-3-1B \citep{gemma2025}. The set includes multilingual and Turkish-specific pre-training, encoder-only, encoder-decoder and decoder-only architectures, and parameter counts from 68M to 1.42B. TurkishBERTweet is pre-trained specifically on 894 million Turkish X posts. We selected model configurations by using grid search over the number of epochs, the learning rate, the batch size and the maximum sequence length, giving 54 configurations, each run under three random seeds. This is 162 runs for each pairing of a model with a dataset and 4,536 fine-tuning runs in total. We used weighted \(F_{1}\) on the held-out test set as the selection criterion in preference to accuracy because of the class imbalance.

\begin{table*}[
    width=.9\linewidth,
    cols=7,
    pos=h,
    align=\centering
]
\centering
\caption{Best single run of the selected classification model for each labelling regime.}
\label{tab:models}
\small
\setlength{\tabcolsep}{5pt}
\begin{tabular}{llrrr}
\hline
Labelling regime & Selected model & Accuracy & Weighted $F_{1}$ & Macro $F_{1}$ \\
\hline
Politics and Government & BERTurk & 93.0 & 92.9 & 86.9 \\
Economy and Finance & ConvBERTurk & 86.0 & 85.6 & 79.7 \\
Media and Society & BERTurk & 86.3 & 86.6 & 80.4 \\
Combined & ConvBERTurk & 88.6 & 88.1 & 81.7 \\
\hline
\end{tabular}
\begin{minipage}{\textwidth}
\vspace{4pt}\footnotesize
\emph{Notes:} All values are percentages on the held-out test set. Figures are the best of three seeded runs and each sits about one seed standard deviation above the corresponding seed average, which is 91.7, 82.8, 84.9 and 87.2 weighted \(F_{1}\) respectively. Macro \(F_{1}\) falls 5.9 to 6.4 points below weighted \(F_{1}\) in all four cases because the neutral class is the smallest in every dataset. Turkish-specific models outperform multilingual pre-training on every dataset.
\end{minipage}
\end{table*}
 
Table~\ref{tab:models} shows the performance of the best models for each of the four datasets (see Table \ref{table:model_comparison} in Appendix for a comparison of all models). We observe that the Politics and Government model is the strongest of the four, consistent with the more explicit sentiment expression of political discourse but also with its heavier class imbalance. When measured against a majority-class baseline the ranking in fact reverses, as Economy and Finance gainied the most accuracy over its own baseline. And the Economy and Finance model is the weakest and least stable across random seeds, with a seed-to-seed standard deviation of 2.6 points of weighted \(F_{1}\), against 1.6 for the next least stable dataset and 1.1 and 1.2 for the other two. Classification noise attenuates an observed relationship rather than manufacturing one. This affects the power of the tests reported below rather than their validity.
 
\subsection{Constructing the public mood series}
\label{subsec:moodseries}
 
We label each post with the model selected for its category, and then with the Combined model, which we apply to the entire corpus. This produces four labelling regimes: three domain-specific and one pooled. We analyse the regimes in parallel rather than ensembling them into a single series. Because whether they agree is itself an empirical question. Post-level labels are mapped to \(\{ + 1,0, - 1\}\) and the public mood score for regime \(j\) on day \(t\) is

\begin{equation}
S_{j,t} = \frac{N_{j,t}^{pos} - N_{j,t}^{neg}}{N_{j,t}},
\label{eq:mood}
\end{equation}

where \(N_{j,t}^{pos}\) and \(N_{j,t}^{neg}\) are the numbers of positive and negative posts assigned to day \(t\) under regime \(j\) and \(N_{j,t}\) is the total number of posts assigned to that day. Neutral posts do not move the score as its factor is 0 but still enter the denominator, and the score is bounded in \(\lbrack - 1,1\rbrack\). Equivalently, \(S_{j,t}\) is the mean of the mapped post-level labels for that day. Weekly and monthly series average over Monday-to-Friday windows and calendar months, giving 500 daily, 104 weekly and 24 monthly observations. The score measures polarity and carries no information about post volume.
 
\subsection{Market data and sub-periods}
\label{subsec:marketdata}
 
We obtained open, close, high, low and volume series for the BIST100 and BIST30 from Yahoo Finance, and the official realised volatility series at 21, 42, 63, 126 and 252-day horizons from Borsa Istanbul. We analyse the two indices in parallel throughout. BIST30 is a subset of BIST100, so agreement between them is a consistency check rather than independent evidence. Disagreement, however, is informative, and reporting both makes the large-capitalisation composition of any result visible.
 
Derived variables are the daily return, the log return, the intraday change (close relative to open), the overnight gap (open relative to previous close), the intraday range in index points and as a percentage of the open, and rolling standard deviations of the daily return at 5, 10 and 21-day windows. The distinction between the range, which measures the magnitude of a day's movement, and the return, which measures its direction, is central to the results.
 
We partition the window into four sub-periods on the basis of events identified from the macroeconomic and political record rather than from inspection of the results. \textbf{2022 as baseline} (252 trading days) with persistent inflation and lira depreciation but no discrete shock, \textbf{2023 Q1} (61 days) containing the 6 February Kahramanmaraş earthquakes and the subsequent suspension of trading, \textbf{2023 Q2} (59 days) containing the May 2023 elections and the beginning of the reversal from heterodox to orthodox monetary policy, and \textbf{2023 H2} (128 days) combining post-election monetary tightening with a pronounced wave of initial public offerings.
 
\subsection{Empirical strategy}
\label{subsec:strategy}
 
We tested all series for a unit root using the Augmented Dickey--Fuller regression
 
\begin{equation}
\Delta y_{t} = \alpha + \gamma y_{t-1} + \sum_{i=1}^{k}\delta_{i}\,\Delta y_{t-i} + \epsilon_{t},
\label{eq:adf}
\end{equation}
with \(k\) selected by the Akaike information criterion and the null \(H_{0}:\gamma = 0\) of a unit root. Testing precedes estimation and also governs interpretation. We report correlations between two non-stationary series where the analysis produced them, but mark them and do not treat them as evidence. Because two series trending over the same window will correlate whether or not they are related. The BIST100 closed 2021 at 1,857.65 points and 2023 at 7,470.18, so this is not a hypothetical concern.
 
We compute Pearson \(r\) and Spearman \(\rho\) between each mood series and each market variable, at all three frequencies, for both indices, over the full period and within each sub-period. We report Spearman alongside Pearson because it is robust to the non-normality of daily financial series. One property of the volatility variables should be recorded here: the rolling standard deviations and the official BIST series are computed over overlapping windows, so consecutive daily values of a 21-day series share twenty of their twenty-one observations. Where both series entering a correlation are serially dependent, the standard error is understated and the associated \(p\)-value is optimistic. We therefore read the volatility correlations as weaker than their nominal levels suggest, and no claim below rests on a volatility correlation alone.
 
We estimate bivariate vector autoregressions for each mood--market pair,
 
\begin{equation}
Y_{t} = c + \sum_{i=1}^{p}A_{i}Y_{t-i} + \epsilon_{t},
\label{eq:var}
\end{equation}
with \(Y_{t} = \left( S_{j,t},m_{t} \right)'\), lag order selected by AIC, and non-stationary components first-differenced before estimation. We test Granger causality in both directions at lags of 1, 2, 3 and 5 trading days using the standard \(F\) test on restricted and unrestricted residual sums of squares. Impulse response functions trace the propagation of a one-standard-deviation innovation. Residual diagnostics are satisfactory throughout. Durbin--Watson statistics of the 56 fitted daily equations run from 1.983 to 2.024, and AIC selects lag orders of five or six for the daily systems.
 
We interpret these tests narrowly, and deliberately so. A rejection of the null establishes that past values of one series improve the forecast of the other beyond that series' own history. It does not establish that one series moves the other. Two features of our design make the weaker reading the only defensible one. A substantial share of Economy and Finance posts are institutional announcements of events the market is simultaneously repricing, and a three-to-five day lag is at least as consistent with gradual information diffusion as with sentiment transmission. Both are taken up in Section~\ref{sec:discussion}. Therefore we report results as predictive precedence and association throughout, never as effect or impact.
 
Our design requires a large number of tests. Counting the mood score alone, which is the measure carried into the causality analyses, and pooling the two indices, there are 576 full-period and 1,728 sub-period correlations and 896 full-period and 2,240 sub-period Granger tests, a total of 5,440. At the 0.05 level, 780 are significant against 272 expected under a global null. The tests are not independent, since the four regimes label the same corpus, the two indices share constituents, and lags of the same variable pair are nested. We apply no formal correction. Instead, we give weight to findings that replicate across labelling regimes, across the two indices and across frequencies, and withhold it from isolated results. This is a criterion for reading the evidence rather than a correction to the individual \(p\)-values. The counts above should be borne in mind wherever a single significant test is reported.
 
Finally, we apply the correlation analysis at the level of individual accounts. An account entered this analysis where its posting provided sufficient trading-day coverage to support the test, which 169 of the 176 accounts satisfied. We report account-level results in aggregate and do not identify any individual account.

\section{Empirical Results}
\label{sec:results}
 
\subsection{Stationarity}
\label{subsec:stationarity}
 
Table~\ref{tab:adf} reports the Augmented Dickey--Fuller tests for both indices. Price levels and the official volatility series at horizons of 42 days and longer contain unit roots on both indices. The returns, the intraday change, the range as a percentage of the open, short-horizon calculated volatility and the official 21-day series are stationary on both.
 
\begin{table*}[
    width=.9\linewidth,
    cols=9,
    pos=h,
    align=\centering
]
\centering
\caption{Augmented Dickey--Fuller tests, daily series, full period.}
\label{tab:adf}
\small
\setlength{\tabcolsep}{5pt}
\begin{tabular}{lrrcrrc}
\hline
 & \multicolumn{3}{c}{BIST100} & \multicolumn{3}{c}{BIST30} \\
\cline{2-4}\cline{5-7}
Series & ADF & $p$ & Class & ADF & $p$ & Class \\
\hline
Open & $-$0.635 & 0.8629 & NS & $-$0.632 & 0.8635 & NS \\
Close & $-$0.719 & 0.8417 & NS & $-$0.711 & 0.8438 & NS \\
Volume & $-$2.670 & 0.0794 & NS & $-$3.430 & 0.0100 & S \\
Return & $-$9.755 & 0.0000 & S & $-$10.000 & 0.0000 & S \\
Intraday change \% & $-$10.033 & 0.0000 & S & $-$10.220 & 0.0000 & S \\
Range & $-$3.098 & 0.0267 & S & $-$2.510 & 0.1130 & NS \\
Range \% & $-$7.932 & 0.0000 & S & $-$8.130 & 0.0000 & S \\
Volatility 5d & $-$6.979 & 0.0000 & S & $-$6.938 & 0.0000 & S \\
Official vol.~21d & $-$3.467 & 0.0089 & S & $-$3.576 & 0.0062 & S \\
Official vol.~42d & $-$2.539 & 0.1063 & NS & $-$2.526 & 0.1092 & NS \\
Official vol.~252d & $-$0.613 & 0.8680 & NS & $-$1.213 & 0.6679 & NS \\
\hline
\end{tabular}
\begin{minipage}{\textwidth}
\vspace{4pt}\footnotesize
\emph{Notes:} S = stationary, NS = non-stationary at the 5\% level. The null is a unit root; augmenting lag order selected by AIC. The log return and the 63 and 126-day official series behave as their neighbours in the table and are omitted for space.
\end{minipage}
\end{table*}
 
Two variables are classified differently across the two indices, and both matter for what follows. The intraday range is stationary on the BIST100 (\(p = 0.0267\)) but not on the BIST30 (\(p = 0.1130\)). Trading volume is the reverse (\(p = 0.0794\) and \(p = 0.0100\)). Both are borderline cases rather than clear rejections, and the range is borderline within the sample as well. It's non-stationary in the 2022 sub-period and stationary in all three 2023 sub-periods. Therefore we read the range results below as robust on the BIST100 and provisional on the BIST30, and the volume results the other way round.
 
\subsection{Contemporaneous association over the full period}
\label{subsec:contemporaneous}
 
Table~\ref{tab:corr} reports daily correlations between each mood series and the market variables for both indices.
 
\begin{table*}[
    width=.9\linewidth,
    cols=10,
    pos=h,
    align=\centering
]
\centering
\caption{Daily correlations between public mood and market variables, full period.}
\label{tab:corr}
\scriptsize
\setlength{\tabcolsep}{2pt}
\begin{tabular}{lrrrrrrrr}
\hline
 & \multicolumn{4}{c}{BIST100} & \multicolumn{4}{c}{BIST30} \\
\cline{2-5}\cline{6-9}
Variable & Pol. & Econ. & Media & Comb. & Pol. & Econ. & Media & Comb. \\
\hline
Close\textsuperscript{\ddag} & 0.093\textsuperscript{*} & 0.085\textsuperscript{\dag} & 0.328\textsuperscript{***} & 0.347\textsuperscript{***} & 0.098\textsuperscript{*} & 0.076\textsuperscript{\dag} & 0.338\textsuperscript{***} & 0.355\textsuperscript{***} \\
Volume\textsuperscript{\S} & $-$0.095\textsuperscript{*} & $-$0.027 & 0.205\textsuperscript{***} & 0.079\textsuperscript{\dag} & $-$0.081\textsuperscript{\dag} & $-$0.033 & 0.139\textsuperscript{**} & 0.038 \\
Return & $-$0.007 & 0.004 & 0.044 & 0.047 & $-$0.014 & $-$0.005 & 0.040 & 0.045 \\
Intraday change \% & 0.010 & $-$0.020 & 0.016 & 0.020 & 0.005 & $-$0.029 & 0.019 & 0.023 \\
Range\textsuperscript{\S} & 0.029 & $-$0.010 & 0.257\textsuperscript{***} & 0.216\textsuperscript{***} & 0.039 & $-$0.024 & 0.289\textsuperscript{***} & 0.239\textsuperscript{***} \\
Range \% & $-$0.024 & $-$0.088\textsuperscript{*} & 0.120\textsuperscript{**} & 0.065 & $-$0.022 & $-$0.102\textsuperscript{*} & 0.128\textsuperscript{**} & 0.064 \\
Volatility 21d (rolling) & 0.109\textsuperscript{*} & $-$0.018 & 0.170\textsuperscript{***} & 0.167\textsuperscript{***} & 0.106\textsuperscript{*} & $-$0.045 & 0.214\textsuperscript{***} & 0.196\textsuperscript{***} \\
Official vol.~252d\textsuperscript{\ddag} & 0.164\textsuperscript{***} & $-$0.011 & 0.354\textsuperscript{***} & 0.354\textsuperscript{***} & 0.181\textsuperscript{***} & $-$0.031 & 0.439\textsuperscript{***} & 0.422\textsuperscript{***} \\
\hline
\end{tabular}
\begin{minipage}{\textwidth}
\vspace{4pt}\footnotesize
\emph{Notes:} Pearson \(r\) between the daily public mood score of each labelling regime and the market variable named. Pol.~= Politics and Government, Econ.~= Economy and Finance, Media = Media and Society, Comb.~= Combined. \(n = 500\) on the BIST100 and 499 on the BIST30 for the close, volume, intraday change and range; 499 and 498 for the return and the official 252-day series; and 479 and 478 for the 21-day rolling measure. \(^{\dagger}\, p < 0.10\), \(^{*}\, p < 0.05\), \(^{**}\, p < 0.01\), \(^{***}\, p < 0.001\). \textsuperscript{\ddag} marks a series that is non-stationary on both indices; \textsuperscript{\S} marks one whose classification differs between them.
\end{minipage}
\end{table*}
There is no association between public mood and the direction of the daily return. The largest of the eight return coefficients across the two indices is 0.047 and none are significant. The same holds for the intraday change and for the overnight gap. Whatever these accounts are doing, they are not moving with the sign of the daily price change.
 
The largest coefficients in the table sit on the closing level and on the 252-day official volatility. Both series carry unit roots on both indices. A coefficient of 0.347 between Combined mood and the closing level of an index that quadrupled over the sample indicates that two series trended together. We report these rows because the analysis produced them and omitting them would misrepresent the output, but they are not evidence of a relationship.
 
What survives the stationarity filter is the magnitude of the daily move. Media and Society mood correlates with the intraday range at \(r = 0.257\) on the BIST100 and \(r = 0.289\) on the BIST30, and Combined at \(r = 0.216\) and \(r = 0.239\), all at \(p < 0.001\). The 21-day calculated volatility, which is stationary on both indices, behaves identically at \(r = 0.170\) and \(r = 0.214\) for Media and Society. Days on which the index travelled through a wide intraday band were days on which Media and Society and Combined mood ran higher. Rank coefficients exceed the linear ones, so the association is monotonic rather than linear. As noted above, the range series itself is borderline stationary. It fails the test on the BIST30, so the 21-day volatility results carry the more secure version of this finding.
 
The Economy and Finance regime is the exception and runs the other way. Its coefficients against every volatility measure are negative on both indices. Among the variables shown, the only one reaching significance is the range as a percentage of the open, at $-$0.088 on the BIST100 and $-$0.102 on the BIST30. The change in trading volume, not tabulated, is the only other Economy and Finance coefficient to reach the 0.05 level, at $-$0.114 and $-$0.092. Of the four regimes, the one composed of accounts that talk about the economy is the one whose mood moves least, and most negatively, with same-day market activity.
 
\subsection{Frequency aggregation}
\label{subsec:frequency}
 
Table~\ref{tab:freq} repeats the range and return correlations at weekly and monthly aggregation.
 
\begin{table*}[
    width=.9\linewidth,
    cols=9,
    pos=h,
    align=\centering
]
\centering
\caption{Public mood and market correlations across aggregation frequencies.}
\label{tab:freq}
\footnotesize
\setlength{\tabcolsep}{2pt}
\begin{tabular}{lrrrrrrrr}
\hline
 & \multicolumn{4}{c}{BIST100} & \multicolumn{4}{c}{BIST30} \\
\cline{2-5}\cline{6-9}
Variable & Pol. & Econ. & Media & Comb. & Pol. & Econ. & Media & Comb. \\
\hline
Range, daily & 0.029 & $-$0.010 & 0.257\textsuperscript{***} & 0.216\textsuperscript{***} & 0.039 & $-$0.024 & 0.289\textsuperscript{***} & 0.239\textsuperscript{***} \\
Range, weekly & 0.087 & 0.128 & 0.363\textsuperscript{***} & 0.372\textsuperscript{***} & 0.098 & 0.100 & 0.408\textsuperscript{***} & 0.402\textsuperscript{***} \\
Range, monthly & 0.214 & 0.214 & 0.482\textsuperscript{*} & 0.523\textsuperscript{**} & 0.233 & 0.162 & 0.531\textsuperscript{**} & 0.558\textsuperscript{**} \\
Return, daily & $-$0.007 & 0.004 & 0.044 & 0.047 & $-$0.014 & $-$0.005 & 0.040 & 0.045 \\
Return, weekly & 0.160 & $-$0.046 & 0.108 & 0.097 & 0.173\textsuperscript{\dag} & 0.002 & 0.097 & 0.111 \\
Return, monthly & 0.060 & $-$0.468\textsuperscript{*} & 0.263 & 0.055 & 0.061 & $-$0.402\textsuperscript{\dag} & 0.240 & 0.059 \\
\hline
\end{tabular}
\begin{minipage}{\textwidth}
\vspace{4pt}\footnotesize
\emph{Notes:} Pearson \(r\). Daily \(n = 500\) (BIST100) and 499 (BIST30), 499 and 498 for the return; weekly \(n = 104\); monthly \(n = 24\). Significance and abbreviations as in Table~\ref{tab:corr}. Monthly coefficients rest on 24 observations and most monthly market variables fail the stationarity test; they are reported for completeness only.
\end{minipage}
\end{table*}
 
The range relationship strengthens monotonically with aggregation. On the BIST100 the Media and Society coefficient rises from 0.257 daily to 0.363 weekly, and Combined from 0.216 to 0.372. The pattern repeats with BIST30. The return relationship does not strengthen. It remains insignificant at every frequency on both indices, where the largest coefficient is 0.173 at the weekly frequency and significant only at the 0.10 level.
 
The asymmetry is informative. If the daily null on returns reflected insufficient power, aggregation should improve it yet it does not. If public mood relates to market activity through attention and the gradual diffusion of narratives rather than through the pricing of new fundamental information, accumulation over longer windows should strengthen a magnitude relationship while leaving a direction relationship absent. Thus, H3 is supported for magnitude and rejected for direction.
 
\subsection{Predictive precedence}
\label{subsec:precedence}
 
Table~\ref{tab:granger} reports the number of significant Granger tests by direction, regime, index and frequency.
 
\begin{table*}[
    width=.9\linewidth,
    cols=10,
    pos=h,
    align=\centering
]
\centering
\caption{Significant Granger causality tests, full period.}
\label{tab:granger}
\footnotesize
\setlength{\tabcolsep}{3pt}
\begin{tabular}{llrrrrrrrr}
\hline
 & & \multicolumn{4}{c}{Daily} & \multicolumn{4}{c}{Weekly} \\
\cline{3-6}\cline{7-10}
Index & Direction & Pol. & Econ. & Media & Comb. & Pol. & Econ. & Media & Comb. \\
\hline
BIST100 & Mood $\rightarrow$ Market & 1/28 & 11/28 & 1/28 & 2/28 & 0/28 & 7/28 & 0/28 & 0/28 \\
BIST100 & Market $\rightarrow$ Mood & 1/28 & 5/28 & 2/28 & 3/28 & 0/28 & 0/28 & 1/28 & 1/28 \\
BIST30 & Mood $\rightarrow$ Market & 0/28 & 9/28 & 0/28 & 0/28 & 1/28 & 4/28 & 0/28 & 0/28 \\
BIST30 & Market $\rightarrow$ Mood & 1/28 & 3/28 & 0/28 & 0/28 & 0/28 & 0/28 & 1/28 & 1/28 \\
\hline
\end{tabular}
\begin{minipage}{\textwidth}
\vspace{4pt}\footnotesize
\emph{Notes:} Counts of tests significant at the 0.05 level out of the tests run, across seven market variables and four lags (1, 2, 3 and 5 periods). Abbreviations as in Table~\ref{tab:corr}.
\end{minipage}
\end{table*}
 
Predictive precedence is concentrated by domain. Economy and Finance accounts for 11 of the 28 daily mood-to-market tests on the BIST100 and nine of 28 on the BIST30, where chance would place roughly 1.4. These include the daily return at lags of three (\(F = 5.04\), \(p = 0.0019\)) and five (\(F = 3.06\), \(p = 0.0098\)) trading days on the BIST100, replicated on the BIST30 at three (\(p = 0.0054\)) and five (\(p = 0.0167\)), together with the closing level, the intraday change and trading volume. The result survives a change of frequency, seven of 28 weekly tests are significant on the BIST100 and four of 28 on the BIST30. No other regime produces more than two significant mood-to-market tests at either frequency on either index. H4 is supported for this direction, and H2 is supported in the specific sense that the domain determines which market property sentiment relates to.
 
The market-to-mood direction is thinner over the full period but contains the single strongest test in the analysis. BIST100 daily range predicts Combined mood one day ahead at \(F = 14.18\) (\(p = 0.0002\)) with the same relationship appearing in Economy and Finance at \(F = 10.53\) and in Media and Society at \(F = 4.67\). Mood also predicts the range at lag one in those three regimes, which makes the two jointly determined at that horizon rather than one leading the other. This bidirectional lag-one pattern is specific to the BIST100. On the BIST30 the range predicts Economy and Finance mood but mood does not predict the range in any regime.
 
The lag structure is worth stating plainly. Market-to-mood relationships appear at one day lag onward. Mood-to-market relationships appear at three and five days lag. Under the straightforward mechanism in which an influential account posts, investors read it, and the price moves at the next open, a lag-one result is the first thing that should appear. It appears at neither frequency and is absent from the contemporaneous correlations. Impulse responses put the magnitudes in perspective. On BIST100 the largest market-to-mood response is 2.6\% of a mood standard deviation on the day following a one-standard-deviation market innovation.
 
\subsection{State dependence}
\label{subsec:statedependence}
 
Table~\ref{tab:subgranger} reports the sub-period Granger counts by direction.
 
\begin{table*}[
    width=.9\linewidth,
    cols=7,
    pos=h,
    align=\centering
]
\centering
\caption{Significant daily Granger causality tests by sub-period and direction.}
\label{tab:subgranger}
\small
\setlength{\tabcolsep}{5pt}
\begin{tabular}{lrrr}
\hline
Sub-period & Tests per direction & Mood $\rightarrow$ Market & Market $\rightarrow$ Mood \\
\hline
\multicolumn{4}{l}{\emph{Panel A: BIST100}} \\
2022 baseline & 112 & 6 & 0 \\
2023 Q1 earthquake & 112 & 4 & 7 \\
2023 Q2 election & 112 & 4 & 43 \\
2023 H2 IPO boom & 112 & 3 & 10 \\

\multicolumn{4}{l}{\emph{Panel B: BIST30}} \\
2022 baseline & 112 & 5 & 1 \\
2023 Q1 earthquake & 112 & 3 & 5 \\
2023 Q2 election & 112 & 7 & 36 \\
2023 H2 IPO boom & 112 & 0 & 7 \\
\hline
\end{tabular}
\begin{minipage}{\textwidth}
\vspace{4pt}\footnotesize
\emph{Notes:} Counts of tests significant at the 0.05 level, pooled across the four labelling regimes, seven market variables and four lags.
\end{minipage}
\end{table*}
 
In 2022, across 112 daily tests on BIST100, no market variable precedes public mood under any regime. In the election quarter of 2023 the market precedes mood in 43 of 112 tests while mood precedes the market in four. The BIST30 reproduces the reversal at 36 and 7. All four labelling regimes flip together. Politics and Government, which produces a single significant mood-to-market result in the full-period analysis, produces 14 market-to-mood results within that one quarter on the BIST100, covering the opening and closing levels, the return, the intraday change, the range and the volume at lags of two to five days.
 
Table~\ref{tab:subcorr} shows the same instability in the contemporaneous coefficients.
 
\begin{table*}[
    width=.9\linewidth,
    cols=9,
    pos=h,
    align=\centering
]
\centering
\caption{Daily correlations by sub-period.}
\label{tab:subcorr}
\scriptsize
\setlength{\tabcolsep}{2pt}
\begin{tabular}{lrrrrrrrr}
\hline
 & \multicolumn{4}{c}{BIST100} & \multicolumn{4}{c}{BIST30} \\
\cline{2-5}\cline{6-9}
Variable & Pol. & Econ. & Media & Comb. & Pol. & Econ. & Media & Comb. \\
\hline
\multicolumn{9}{l}{\emph{Daily return}} \\
2022 baseline & 0.054 & 0.100 & 0.105\textsuperscript{\dag} & 0.177\textsuperscript{**} & 0.048 & 0.098 & 0.096 & 0.167\textsuperscript{**} \\
2023 Q1 earthquake & 0.070 & 0.013 & $-$0.045 & 0.067 & 0.055 & $-$0.003 & $-$0.046 & 0.064 \\
2023 Q2 election & $-$0.228\textsuperscript{\dag} & $-$0.164 & $-$0.095 & $-$0.281\textsuperscript{*} & $-$0.191 & $-$0.178 & $-$0.061 & $-$0.247\textsuperscript{\dag} \\
2023 H2 IPO boom & $-$0.019 & $-$0.035 & 0.149\textsuperscript{\dag} & 0.118 & $-$0.056 & $-$0.056 & 0.121 & 0.088 \\
 
\multicolumn{9}{l}{\emph{Intraday range}} \\
 
2022 baseline & 0.003 & $-$0.124\textsuperscript{*} & 0.432\textsuperscript{***} & 0.370\textsuperscript{***} & 0.017 & $-$0.134\textsuperscript{*} & 0.450\textsuperscript{***} & 0.389\textsuperscript{***} \\
2023 Q1 earthquake & $-$0.259\textsuperscript{*} & 0.007 & $-$0.039 & $-$0.102 & $-$0.260\textsuperscript{*} & 0.000 & $-$0.044 & $-$0.110 \\
2023 Q2 election & 0.233\textsuperscript{\dag} & $-$0.178 & $-$0.001 & 0.001 & 0.216 & $-$0.178 & 0.002 & $-$0.029 \\
2023 H2 IPO boom & $-$0.080 & $-$0.026 & $-$0.000 & $-$0.156\textsuperscript{\dag} & $-$0.077 & $-$0.047 & 0.049 & $-$0.137 \\
 
\multicolumn{9}{l}{\emph{Trading volume}} \\
 
2022 baseline & 0.051 & 0.018 & 0.425\textsuperscript{***} & 0.441\textsuperscript{***} & 0.079 & 0.018 & 0.302\textsuperscript{***} & 0.336\textsuperscript{***} \\
2023 Q1 earthquake & $-$0.143 & 0.356\textsuperscript{**} & $-$0.240\textsuperscript{\dag} & $-$0.142 & $-$0.159 & 0.356\textsuperscript{**} & $-$0.229\textsuperscript{\dag} & $-$0.148 \\
2023 Q2 election & $-$0.239\textsuperscript{\dag} & $-$0.362\textsuperscript{**} & $-$0.285\textsuperscript{*} & $-$0.483\textsuperscript{***} & $-$0.253\textsuperscript{\dag} & $-$0.335\textsuperscript{**} & $-$0.304\textsuperscript{*} & $-$0.500\textsuperscript{***} \\
2023 H2 IPO boom & 0.004 & $-$0.231\textsuperscript{**} & 0.500\textsuperscript{***} & 0.112 & $-$0.061 & $-$0.261\textsuperscript{**} & 0.476\textsuperscript{***} & 0.064 \\
\hline
\end{tabular}
\begin{minipage}{\textwidth}
\vspace{4pt}\footnotesize
\emph{Notes:} Pearson \(r\). Observations: 2022 baseline 252 (251 for the return), earthquake 61, election 59, IPO boom 128. Significance and abbreviations as in Table~\ref{tab:corr}.
\end{minipage}
\end{table*}
 
The sign reverses between 2022 and the election quarter on both indices at once. Combined mood correlates with the daily return at \(r = 0.177\) (\(p = 0.005\)) in 2022 on the BIST100 and \(r = 0.167\) on the BIST30. In the election quarter the same coefficients are $-$0.281 and $-$0.247. Media and Society mood correlates with the range at 0.432 in 2022 and at essentially zero in the election quarter. All four regimes return negative correlations with trading volume in the election quarter, significant in three of four on both indices, with Combined reaching $-$0.483 and $-$0.500. All four also return a negative and significant correlation with the official 21-day volatility in that quarter on the BIST100, at $-$0.259 in Economy and Finance to $-$0.530 in Combined, which is the only period in the entire correlation analysis where all four regimes agree in sign and significance. A higher mood in the election quarter went together with a calmer and thinner market. This supports H5.
 
The earthquake quarter produces some of the smallest correlation coefficients of the four sub-periods. It is worth stating because the literature on sentiment and crisis would predict the opposite. Three of the twelve BIST100 cells in Table~\ref{tab:subcorr} reach significance on 61 observations, and each carries the opposite sign to the full-period coefficient of the same regime on the same variable.
 
\subsection{Regime disagreement}
\label{subsec:regimedisagreement}
 
The four labelling regimes do not measure the same quantity, and the Combined regime is not an average of the other three. Its mean daily mood score of 0.016 sits against a Politics and Government mean of 0.510, a Media and Society mean of $-$0.002, an Economy and Finance mean of $-$0.093. The unweighted three-category mean is 0.138. The same ordering holds within all four sub-periods. In every one, the Combined series falls within 0.023 of the largest contributing category and never closer than 0.094 to the average of the three. Media and Society covers 297,857 of the 610,422 posts, and the pooled series reproduces it.
 
The impact is visible throughout the tables above. Combined tracks Media and Society on the range and volatility results and departs from it elsewhere, across the thirteen daily market variables the two agree in sign on twelve. The Economy and Finance predictive results have a much weaker counterpart in the Combined series, which produces two significant daily mood-to-market tests against eleven in the regime it draws from. And the Politics and Government mood score, which is positive on all 500 trading days, has a sign that never varies. So the aggregate series cannot register the shifts in valence that its individual constituents display.
 
\subsection{Account-level analysis}
\label{subsec:accountlevel}
 
Aggregating a category into one daily series assumes that its members behave alike. Table~\ref{tab:accounts} reports the correlations we repeated account by account.
 
\begin{table*}[
    width=.9\linewidth,
    cols=8,
    pos=h,
    align=\centering
]
\centering
\caption{Accounts correlating significantly with selected market variables, by category.}
\label{tab:accounts}
\scriptsize
\setlength{\tabcolsep}{3pt}
\begin{tabular}{lrrrrrrr}
\hline
 & & \multicolumn{3}{c}{BIST100} & \multicolumn{3}{c}{BIST30} \\
\cline{3-5}\cline{6-8}
Category & Accounts & Return & Range \% & Vol 21d & Return & Range \% & Vol 21d \\
\hline
Politics and Government & 70 & 1 (1.4) & 3 (4.3) & 17 (24.3) & 1 (1.4) & 5 (7.1) & 15 (21.4) \\
Economy and Finance & 67 & 11 (16.4) & 12 (17.9) & 8 (11.9) & 11 (16.4) & 13 (19.4) & 8 (11.9) \\
Media and Society & 64 & 3 (4.7) & 7 (10.9) & 15 (23.4) & 3 (4.7) & 6 (9.4) & 14 (21.9) \\
All accounts pooled & 169 & 12 (7.1) & 22 (13.0) & 33 (19.5) & 11 (6.5) & 21 (12.4) & 33 (19.5) \\
\hline
\end{tabular}
\begin{minipage}{\textwidth}
\vspace{4pt}\footnotesize
\emph{Notes:} Number of individual accounts whose daily mood series correlates with the named variable at the 0.05 level, with the percentage of the category in parentheses. Chance alone would place roughly 5\% in each column. The first three rows use the category model on the accounts of that category; the final row applies the Combined model to all accounts with sufficient coverage and is therefore not the sum of the three, which in any case exceeds 169 because the category assignment is not disjoint. No individual account is identified.
\end{minipage}
\end{table*}
 
Disaggregation reproduces the aggregate picture on the return and reverses it on volatility. Eleven of the 67 Economy and Finance accounts (16.4\%) correlate significantly with the daily return on both indices, against one of 70 in Politics and Government and three of 64 in Media and Society. The first figure is well above chance and the second well below it. The signs of those eleven are split six positive to five negative. So the category result is a matter of individual accounts relating to the index rather than of the category moving with it in a common direction.
 
On volatility the ordering reverses. Seventeen Politics and Government accounts (24.3\%) correlate significantly with the official 21-day volatility series, twelve of them positively. It is the largest share of any category and more than twice the 11.9\% recorded by Economy and Finance. BIST30 gives 21.4\% and 11.9\%. The largest single account-level correlation with the daily return found anywhere in the analysis (\(|r| = 0.305\)) also belongs to a Politics and Government account, although on the trading-day coverage available to that account it stops just short of significance (\(p = 0.0625\)).
 
Individual political accounts do relate to Borsa Istanbul, and more often than the accounts of the other two categories. What the category series loses, it loses in the aggregation. A few dominant accounts drowning out the rest would be the obvious explanation. But Politics and Government is the least concentrated of the three categories, its top decile of accounts holding 28.9\% of the category's posts against 44.0\% in Economy and Finance and 54.8\% in Media and Society. Institutional register runs evenly through the whole account list, so averaging over it removes the variation that the individual accounts carry.

\section{Discussion}
\label{sec:discussion}
 
\subsection{What each domain measures}
\label{subsec:whatdomains}
 
The results support H2 in a specific form. The three domains do not differ in the strength of a common relationship. They differ in which property of the market they attach to. Media and Society mood, and the Combined series that inherits its properties, relates to the magnitude of daily price movement. Economy and Finance mood relates to almost nothing contemporaneously and carries the predictive precedence, at lags of three days and longer. Politics and Government mood, in aggregate, relates to little at the category level beyond the volatility series, while its individual members relate to volatility more often than those of any other category.
 
This ordering follows from the composition of the account sets. Media and Society is the group least constrained by office and broadest in subject matter. As such its aggregate tone comes closest to a general public mood. Its full-period mean of $-$0.002 is the behaviour of a series responding to events in both directions. Economy and Finance sits nearest the flow of financial and policy information, its mean of $-$0.093 remains negative in all four sub-periods. So accounts covering the Turkish economy described it in negative terms across two years in which the index quadrupled. Politics and Government is constrained by office to communicate positively, with a mean of 0.510 and no negative day in 500. Three different quantities are being measured under one label, and the domain-specific design is what makes them separable.
 
\subsection{Magnitude without direction}
\label{subsec:magnitude}
 
Our most consistent result is also the one that constrains interpretation most tightly, and it is what H1 predicted in its semi-strong form. Publicly available tone carries no information about the direction of the next price change, only about how far the index travelled within the day. Two features point away from information pricing, towards attention. The relationship strengthens monotonically with aggregation while the return relationship does not. The daily null on returns is not a power problem. And the rank coefficients exceed the linear ones throughout, so extreme days matter more than a linear measure suggests. This is consistent with the attention and gradual-diffusion arguments in Section~\ref{sec:theory}, salient content is more likely to enter the decisions of investors who cannot process all available signals \citep{barber2008}, and information reaching investors at different speeds produces effects that accumulate over horizons rather than appearing at once \citep{hong1999}. Under that reading, a wide-range day and elevated public discussion are both expressions of a day on which something happened.
 
\subsection{The direction and horizon of transmission}
\label{subsec:transmission}
 
Both directions appear, and the lag structure separating them is the strongest internal evidence against a simple sentiment-transmission story. Three to five trading days is a long interval for a mood to travel but a comfortable one for information to diffuse outward from an announcement through a market in which a growing share of participants were new. That direction is not fixed, this was already visible in the Turkish evidence. \citet{ates2021} found returns preceding sentiment over their long sample and sentiment preceding returns only inside a politically eventful window.
 
The Economy and Finance predictive results are the strongest in the study and also the most exposed. That category contains the Central Bank, the Capital Markets Board, Borsa Istanbul and the economic ministries, so a large share of its posts are announcements rather than opinions. When a rate decision, an inflation print, a disclosure or a regulatory measure enters the mood score as a positively or negatively toned post on the day it appears, the market reprices over the following days. Granger test can separate that sequence from mood carrying information. This is not a hypothetical omitted variable but a documented property of the account list, and it is why we report the finding as predictive precedence rather than as an effect.
 
One feature cuts the other way. The Economy and Finance labels come from the model with the largest seed-to-seed variance of the four. Classification noise attenuates a relationship rather than manufacturing one, so a noisier series producing a stronger result is weaker evidence in favor of the relationship. The finding rests nonetheless on the least stable measurement in the study.
 
\subsection{State dependence}
\label{subsec:statedependence2}
 
H5 is the hypothesis the data support most strongly. It is supported on both indices simultaneously and under all four regimes. The pattern is what the Adaptive Markets Hypothesis anticipates \citep{lo2004}, a relationship observed under one set of conditions need not persist when the composition of participants, the credibility of public messages and the decision rules in use change together. It is also what the political-uncertainty literature would predict. When the resolution of political uncertainty is itself the dominant national news, commentary that follows prices rather than preceding them is close to what should be expected \citep{pastor2013}. The corpus registers the same shift internally. Politics and Government mood rises from 0.482 in 2022 to 0.637 in the election quarter, the largest sub-period movement of any regime, while Economy and Finance falls to $-$0.121. The divergence between political messaging and economic commentary peaks where the macroeconomic record would put it.
 
The earthquake quarter requires separate comment because it produces the least of the four sub-periods, the opposite of what the crisis-sentiment literature would predict. Three explanations are available and they are not seperate. Trading was suspended for part of the quarter, and the missing days immediately after 6 February are precisely where a leading relationship would have to appear. The posts of these accounts during a humanitarian catastrophe are about the catastrophe, so for several weeks the series may not be measuring economic mood at all. And 61 observations is a thin basis for a lag-five test. The first two explanations are the more consequential, because both imply that the measure stops capturing the same construct during exactly the events where it is expected to matter most.
 
\subsection{Aggregation as a measurement problem}
\label{subsec:aggregation}

The comparison across labelling regimes yields a result that generalises beyond this dataset. A sentiment score computed over all posts in a day does not represent the corpus. It represents whichever category posted most. The Politics and Government series shows the same issue more starkly, averaging 70 accounts constrained by office to communicate positively produces a series whose sign never changes in 500 trading days. So whatever valence its members express individually is compressed into a narrow positive band around a mean of 0.510. The account-level results in Table~\ref{tab:accounts} confirm that this is a property of the method rather than of the accounts. The individual members of that category relate to volatility more often than those of any other.
 
The weakness of the aggregate Politics and Government results is therefore a limitation of the aggregation and not evidence that political speech is unrelated to Borsa Istanbul. The artefact is available to any study that scores an institutionally constrained account set daily and then averages, nothing in the output of a daily mean announces its presence.
 
\subsection{Alternative explanations}
\label{subsec:alternatives}
 
For attributing market movements to a single driver, Turkiye’s January 2022-December 2023 period is unsuitable: it combines recurring lira depreciation, inflation above 80\%, a policy-rate cut to 8.5\% followed by post-May 2023 tightening, the February 2023 Kahramanmaraş earthquakes, and an IPO surge that altered the investor base.
 
Each has a documented impact on the index. Changepoint analysis of Turkish equity volatility relates shifts to central bank decisions, domestic political incidents and global shocks \citep{altinbas2025}. Over a longer sample, inflation and inflation uncertainty are both found to raise BIST100 returns, consistent with Turkish equities being held as an inflation hedge \citep{esen2025}. Elections are known to produce unusual returns on Borsa Istanbul accumulating from roughly a month before the vote to two weeks after it, with volatility running one and a half to two times its pre-election average \citep{kayacetin2023}. An event study of the 2023 election in particular reports strongly positive cumulative abnormal returns around the second round \citep{bash2023}. For the earthquake, a Bayesian structural time series analysis using global indices as controls estimates a substantial negative effect on the BIST100 against its counterfactual \citep{khan2024}. Mechanisms unrelated to any account's posts explain the level path of the index, which is the light in which the non-stationary coefficients of Section~\ref{sec:results} should be read.
 
Three confounds affect the significant results. The magnitude relationship may reflect common causes: major events move markets and prompt related posting, which correlation cannot disentangle. The Economy and Finance predictive results may reflect the announcement channel, while the election-quarter reversal coincides with the election, a policy transition, and the start of rate hikes within only 59 trading days. Finally, the polarity-based measure omits posting volume, which other markets suggest can strongly predict volatility \citep{hamraoui2022}.
 
 
 

\section{Conclusion}
\label{sec:conclusion}
 
We examined whether the tone of communication by influential Turkish public accounts has a measurable relationship with Borsa Istanbul. We analysed 610,422 posts published by 176 curated accounts between January 2022 and December 2023, labelled them with four fine-tuned Turkish transformer models under three domain-specific regimes and one pooled regime, then tested the resulting series against the BIST100 and BIST30 over 500 trading days.
 
Five findings hold. Public mood is not associated with the direction of returns under any regime, on either index or at any frequency. It is associated instead with the magnitude of daily price movement, and that association strengthens with aggregation in a manner consistent with attention and gradual diffusion rather than with the pricing of new information. Predictive precedence runs in both directions and is concentrated by domain, mood-to-market at three and five days and market-to-mood at one. The relationship reverses in sign and direction during the 2023 election quarter under all four regimes and on both indices. And the four labelling regimes do not agree, and the pooled series reproduce whichever category contributes the most posts.
 
\subsection{Implications}
\label{subsec:implications}
For regulators and exchange operators, the state dependence result is most relevant. Domain-specific mood indicators appear more useful for identifying episodes when public discourse and market activity become tightly coupled than for predicting market direction. Here, coupling was strongest around a national election, and the market-to-mood direction dominated. This suggests that online commentary in such episodes may reflect price and volatility rather than independently drive instability.

For market participants, the absence of an association with returns applies only to this corpus and construction. Speaker-based sampling and daily averaging discard financial-content selection, posting volume, and within-day timing. The results therefore do not rule out discourse-return relationships; rather, observed associations concern movement magnitude rather than direction, with signs varying across the two-year window.

For researchers, the aggregation result is most transferable. Pooling heterogeneous speaker groups can produce a series whose variance and findings are dominated by the largest group. Category-level and account-level analyses should therefore precede interpretation of pooled measures.
 
\subsection{Limitations}
\label{subsec:limitations}
 
Five limitations bound our results. We measure public mood rather than investor sentiment, and the one-million-follower threshold ensures reach but not investor relevance. The two-year window includes major shocks—a currency crisis, inflation above 80\%, a monetary-regime change, a catastrophic earthquake, and a national election—and the results show that relationships were not stable within it, limiting extrapolation. Sub-periods contain only 59-128 trading days, adequate for correlations but limited for five-lag Granger tests; stationarity classifications also differ across periods, making the specifications non-identical. The Economy and Finance model is the weakest and least stable across seeds, yet underlies the strongest predictive result. We conduct many tests without formal multiple-comparison correction, so isolated significant findings warrant caution. Our main robustness criterion is replication across regimes, indices, and frequencies.
 
\subsection{Further research}
\label{subsec:further}
 
Three extensions follow. Adding posting volume alongside polarity would separate salience from tone and test whether volume better predicts market magnitude. Time-varying or recursive-window causality over a longer sample could assess whether the election-quarter market-to-mood coupling recurs in similar episodes. Finally, replication on Turkish investor forums or the same accounts during a calmer period would help establish which findings are robust beyond this dataset.

\printcredits

\bibliographystyle{cas-model2-names}

\bibliography{cas-refs}


\clearpage

\appendix
\label{sec:app}

\section{Performance of the Seven Fine-tuned Models on the Politics and Government Dataset}

\renewcommand{\thetable}{\Alph{section}.\arabic{table}}
\setcounter{table}{0}

Table \ref{table:model_comparison} demonstrates the performance of the seven different fine-tuned models on the Politics and Government dataset.
\begin{table}
\centering
\setlength{\tabcolsep}{3pt}
\caption[Performance of the seven fine-tuned models on the Politics and Government dataset]{Best single run of each of the seven fine-tuned models on the Politics and Government dataset, sorted by weighted F1. All values are percentages.}
\vskip\baselineskip
\footnotesize
\begin{tabular}{lcccc}
\hline
\textbf{Model} & \textbf{Accuracy} & \textbf{Weighted F1} & \textbf{Precision} & \textbf{Recall} \\\hline
BERTurk & 93.0 & 92.9 & 93.0 & 93.0 \\
ConvBERTurk & 92.3 & 92.1 & 92.5 & 92.3 \\
DistilBERTurk & 89.0 & 88.7 & 89.0 & 89.0 \\
TurkishBERTweet & 88.7 & 88.3 & 88.4 & 88.7 \\
mBERT & 87.0 & 86.8 & 87.0 & 87.0 \\
Gemma-3-1B & 87.3 & 86.6 & 87.0 & 87.3 \\
BERT5urk & 83.7 & 80.8 & 78.1 & 83.7 \\
\hline
\end{tabular}
\label{table:model_comparison}
\end{table}

\end{document}